\documentclass[onecolumn]{IEEEtran}
\usepackage{epsfig,graphicx,verbatim,subfigure,multirow,stackengine,placeins}
\graphicspath{ {images/} }
\def\delequal{\mathrel{\ensurestackMath{\stackon[1pt]{=}{\scriptstyle\Delta}}}}
\begin{document}
\title{Feature Fusion for Robust Patch Matching\\With Compact Binary Descriptors}
%
%
\author{\IEEEauthorblockN{Andrea Migliorati\IEEEauthorrefmark{1},
Attilio Fiandrotti\IEEEauthorrefmark{2},
Gianluca Francini\IEEEauthorrefmark{3}, 
Skjalg Lepsoy\IEEEauthorrefmark{3} and
Riccardo Leonardi\IEEEauthorrefmark{1}}
\IEEEauthorblockA{\IEEEauthorrefmark{1}Information Engineering Department, Universit\`a degli Studi di Brescia, \textit{NAME.SURNAME}@unibs.it}
\IEEEauthorblockA{\IEEEauthorrefmark{2}Politecnico di Torino, Turin, Italy, attilio.fiandrotti@polito.it}
\IEEEauthorblockA{\IEEEauthorrefmark{3}Telecom Italia S.P.A., Turin, Italy, \textit{NAME.SURNAME}@telecomitalia.it}}
\maketitle
\begin{abstract}
This work addresses the problem of learning compact yet discriminative patch descriptors within a deep learning framework. We observe that features extracted by convolutional layers in the pixel domain are largely complementary to features extracted in a transformed domain. We propose a convolutional network framework for learning binary patch descriptors where pixel domain features are fused with features extracted from the transformed domain. In our framework, while convolutional and transformed features are distinctly extracted, they are fused and provided to a single classifier which thus jointly operates on convolutional and transformed features. We experiment at matching patches from three different datasets, showing that our feature fusion approach outperforms multiple state-of-the-art approaches in terms of accuracy, rate, and complexity. 
\end{abstract}
\IEEEpeerreviewmaketitle
%
%
\section{Introduction}
Comparing image patches is a key task in a number of computer vision applications such as object identification, photo stitching, stereo baseline estimation, etc. Patch comparison usually takes place by comparing local descriptors that are robust to rotations, scale, illumination and, to some extents, perspective, and changes. Descriptors such as SIFT had to be handcrafted according to the specific task or considered signal. Recent advances in deep learning showed that it is possible to train a neural network to automatically learn and compare local descriptors without necessarily resorting to handcrafted descriptors.

Over the past years, a number of descriptors designs based upon Convolutional Neural Networks (CNN) have been proposed \cite{zheng2017sift}. The work of \cite{zagoruyko2015learning} explores different CNN architectures, showing that best results are achieved when pairs of patches are jointly encoded and a decision network is trained to learn an appropriate inter-patch distance metric. The authors of \cite{simo2015discriminative} address the problem of generating SIFT-like descriptors within a deep learning framework, showing that deep learning generated descriptors can be used as a drop-in replacement for SIFT descriptors as they retain key properties such as invariance to rotations, illumination and perspective changes.  In \cite{zbontar2016stereo}, the specific problem of matching wide baseline stereo images by comparing local patches using a siamese CNN is addressed. The authors show that the network patch matching accuracy can be greatly enhanced augmenting the training set of patches by rotation and illumination changes. In \cite{yang2017deepcd}, an approach based on fusing two complementary and asymmetric descriptors extracted from the convolutional domain is proposed. The authors of \cite{zhu2016deep} address the related yet different problem of learning global hash codes over whole images. Their approach is limited to small size images due to complexity constraints. In addition, global descriptors are inherently unsuited for geometric verification, showing the advantages of local descriptors based approaches. Finally, in \cite{yi2016lift} a complete image matching pipeline based on a deep learning framework is presented: whereas such architecture goes beyond the scope of the present work, it clearly shows the potentials of image matching architectures based on deep learning frameworks. 

Our goal is to learn patch descriptors that are highly discriminative and, at the same time, rate efficient and computationally lightweight to generate and compare. According to such requirements, many of the above designs fall short in one or more aspect. In \cite{simo2015discriminative}, descriptors are devised as a drop-in replacement for SIFT descriptors, thus they are encoded as real-valued vectors, whereas binary vectors are desirable in reason of their improved rate efficiency. In \cite{zagoruyko2015learning} best performance is achieved by jointly encoding pairs of patches, an approach that is not suitable for the common case where the reference patch is available only through its descriptor at query time. Such approach also requires replicating the same decision network learned at training time when deployed on the field, impacting significantly in terms of complexity. The approach of \cite{yang2017deepcd} shows state-of-the-art results, however at the expense of duplicating the feature extraction pipeline complexity. For such reasons, how to learn and compare patch descriptors that are at the same time discriminative and efficient both in terms of rate and complexity is still an open research issue.

In this work, we introduce the idea of fusing features from the convolutional layers and from the discrete cosine transform to learn binary patch descriptors within a deep learning framework. DCT features have been previously considered for feature dimensionality and redundancy reduction in patch matching and face recognition problems \cite{sorwar2001texture,pan2000image}. However, to the best of our knowledge, no previous work has considered the problem of fusing features from the convolutional and the DCT domains within a deep learning framework. We propose a framework that is designed around a Siamese network, which allows to independently generate descriptors from single patches. We learn real-valued compact descriptors by minimizing a loss function of the cosine distance between pairs of descriptors, where the cosine distance happens to be the best-fitting real-valued relaxation for the Hamming distance, enabling straightforward descriptors quantization and subsequent comparison. We experiment with three challenging datasets, showing better performance than state-of-the-art competitors for identical bitrate and comparable performance to more complex approaches despite lower bitrate. The rest of this paper is organized as follows. In Sec.~\ref{sec:background} relevant background literature is surveyed. Sec.~\ref{sec:proposed} describes the proposed convolutional approach to patch matching via feature fusion in terms of network architecture and training procedures. Sec.~\ref{sec:experiments} provides a thorough experimental validation of the framework over three sets of real image patches. Sec.~\ref{sec:conclusion} draws the conclusions and discusses potential directions for further investigations.
\section{Background}
\label{sec:background}
This section provides the relevant background on Siamese neural networks, a class of feed-forward artificial neural networks. Siamese networks find application in a number of problems where it is required to learn a similarity or distance function between pairs of equally dimensioned signals, ranging from face verification \cite{chopra2005learning} to real-time object tracking \cite{bertinetto2016fully}. Siamese networks are composed by (at least) two topologically identical and independent \textit{subnetworks} which share an identical set of learnable parameters. In most image and video applications, each subnetwork typically includes one or more convolutional layers for extracting features from the input signal spatial domain. Optionally, a number of fully connected layers projects such features to a typically lower-dimensional space, yielding a vector of features representing the signal provided as input to the subnetwork. The output of each subnetwork can be seen as a (compact) description of the input signal, which is thus often defined as \textit{descriptor}. Finally, one predefined or learnable distance function measures the similarity between the two descriptors produced by each subnetwork. In this context, the network produces as output a measure of the extent to which two input patches are similar or dissimilar. Concerning the patch matching problem addressed by this work, Siamese architectures enable to independently compute each patch descriptor, allowing for patch matching with precomputed descriptors. Therefore, in the following, we use a Siamese architecture as the cornerstone of our network, whereas we leave the application of our feature fusion approach to other network architectures for future research.
\section{Proposed Method}
\label{sec:proposed}
In this section, we first describe a Siamese convolutional network architecture suitable for convolutional and DCT feature fusion. Next, we describe a fully-supervised training procedure with the associated loss function. Finally, we briefly describe the descriptor quantization process.

\begin{figure*}[t]
\centering
\includegraphics[width=15cm]{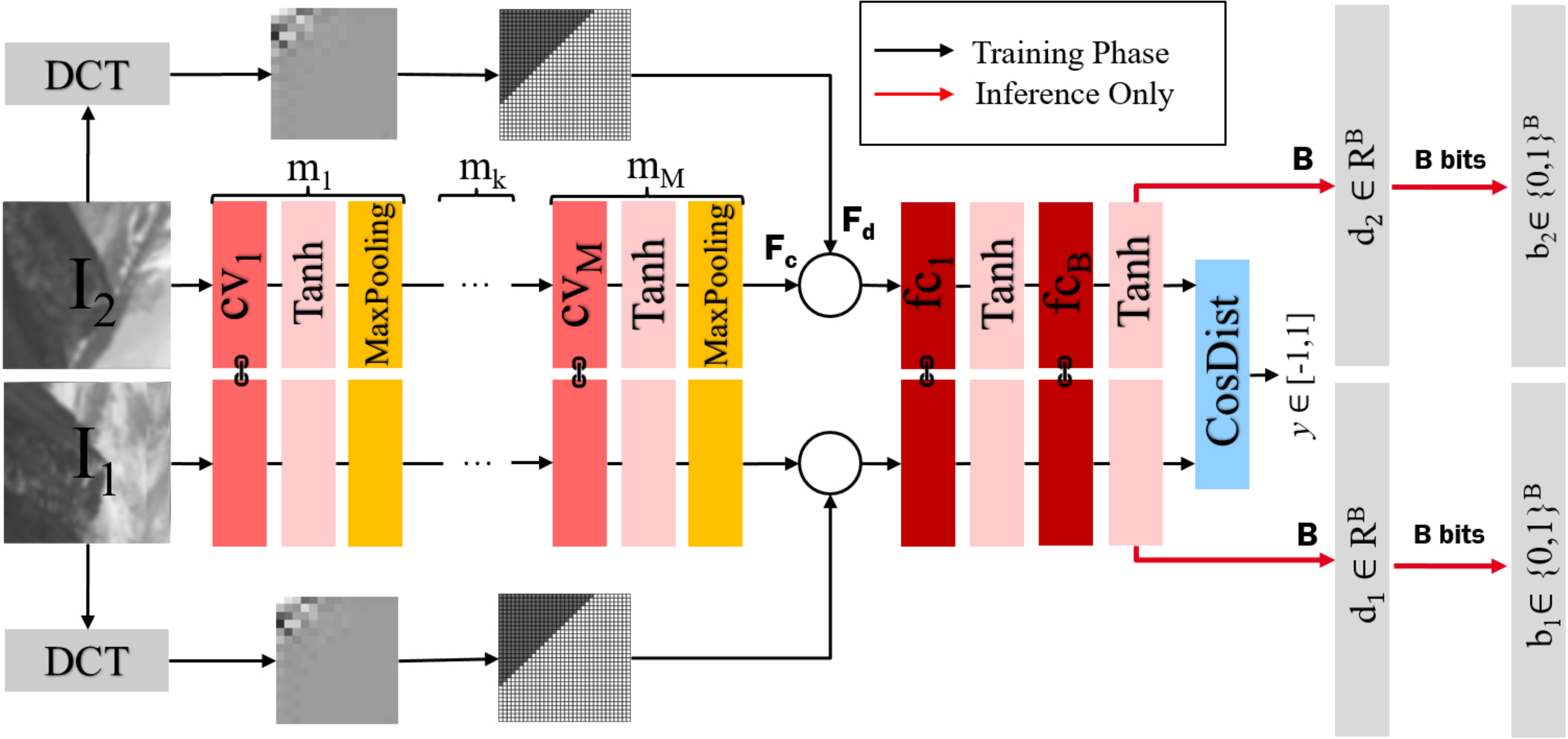}
\caption{Our proposed Siamese convolutional architecture for convolutional and DCT feature fusion.}
\label{fig:struttura}
\end{figure*}

\subsection{Network Architecture}
\label{sec:structure}

The architecture in Fig.~\ref{fig:struttura} relies on a Siamese topology where the two subnetworks receive as input a pair of same size image patches \((I_1,I_2)\).
Each subnetwork is structured as follows: the first part is composed of M \textit{convolutional modules}, where each \textit{k}-th module $m_k$ (\(k \in {1, ..., M} \)) is composed by one convolutional layer with an hyperbolic tangent nonlinearity and one 2$\times$2 max-pooling layer. All the convolutional layers are composed of 5$\times$5 filters (kernels), and convolutions are of the wide type with border padding and one-pixel stride.
The number of filters in the first \((m=1)\) module's convolutional layer is set to \(64\) and doubles at each module, thus the number of \textit{featuremaps} produced as output by each convolutional layer doubles with index \(k\).
However, the resolution of each \textit{featuremap} is reduced by a fourfold factor after each max-pooling layers, so the overall number of features generated by each module shrinks by a factor of two after each \textit{m}-th module.
The number of \textit{convolutional features} produced as output by the \textit{k}-th module is referred to as \(F_C\) in the following.
Fig.~\ref{fig:struttura} shows that, for each subnetwork, a parallel branch of the pipeline implements a 2D discrete cosine transform of the input signal.
Next, a subset of \(F_D\) coefficients from the top-left corner of the transformed patch matrix is selected in zig-zag scanning order starting from the DC coefficient. The result is a number of \(F_D\) additional \textit{DCT features} that complements the \(F_C\) convolutional features.

Finally, the \(F_C\) convolutional features are concatenated with the \(F_D\) DCT features, yielding a vector of \(F_C+F_D\) \textit{fused features} that represents a projection of the input signal to a \((F_C+F_D)\)-dimensional space. Fused features are processed by a first fully connected layer with \(512\) units (neurons) with hyperbolic tangent activation functions. A second fully connected layer with B units, to which we refer as \textit{bottleneck layer} in the following, generates a B-elements patch descriptor. Accordingly, the two subnetworks receive as input the pair of image patches \((I_1,I_2)\) and produce as output the corresponding pair of descriptors \((d_1,d_2)\) of B real-valued elements each. 

In order to establish the similarity between the two descriptors \((d_1,d_2)\), we resort to the cosine distance function, which produces an output in the range [-1,1], and it is defined as 

\begin{equation}
\small
C(d_1,d_2) \delequal \frac{\langle d_1, d_2 \rangle}{\|d_1\|  \hfill \cdot \|d_2\|} = \frac{\sum_{j=1}^{B} d_{1,j} d_{2,j}}{\sqrt{\sum_{j=1}^{B}d_{1,j}^2} \hfill \sqrt{\sum_{j=1}^{B}d_{2,j}^2}}.
\label{eqn:cosine}
\end{equation}
\\*
Such cosine distance represents the ultimate output of the proposed network. Hence, when a pair of patches \(x=(I_1,I_2)\) is fed into the network, the associated cosine distance \(y=C(d_1,d_2)\) between its descriptors is produced. The cosine distance enjoys several useful properties: first, it is always continuous and differentiable, thus it allows fully supervised, end-to-end, training of the network as discussed in the following; moreover, cosine distance affinity with the Hamming distance enables straightforward descriptor quantization and binary comparison as described in Sec.~\ref{sec:quantization}.

Finally, let us define as \textit{network complexity} the number of learnable parameters in the network. Concerning the architecture showed in Fig.~\ref{fig:struttura}, the complexity of the first fully connected layer (\(512\) units) dominates the overall network complexity. Thus, the network complexity is approximately equal to \((F_C +F_D) \cdot 512\), i.e. its complexity increases linearly with the number of fused features.

\subsection{Training}
\label{sec:training}

Let us define as the \textit{i}-th \emph{training sample} the pair of identically sized image patches \(x_i=(I_1^i,I_2^i)\).  \(I_1^i\) and \(I_2^i\) are said to be \textit{matching} (equivalently, that \(x^i\) is a matching sample) whenever \(I_1^i\) and \(I_2^i\) represent the same image detail, and \textit{non-matching} otherwise. Let \(y^i = C(d_1^i, d_2^i)\) and \(l^i\) be the network output and the corresponding target \textit{label} (i.e., the expected network outcome), respectively. Our ultimate goal is to learn network parameters such that the network generates pairs of discriminative descriptors \((d_1^i, d_2^i)\). However, as we aim at training the network end-to-end with a fully supervised approach, we recast the problem on learning parameters such that the network, given as input the \textit{i}-th sample \(x^i\), generates an output for which \(y^i=l^i\). To this end, the choice of the training label \(l^i\) is pivotal. Concerning pairs of matching patches, the network is expected to produce similar descriptors such that \(C(d_1^i, d_2^i) \approx 1\), so we impose \(l^i=1\) for matching samples. Considering the comparison of non-matching pairs of patches, we observe that such process is equivalent to comparing random i.i.d. patches \((I_1^i,I_2^i)\), which in turn is equivalent to measuring the cosine distance between the associated i.i.d. descriptors \((d_1^i,d_2^i)\). Given two zero-mean i.i.d. \((d_1^i,d_2^i)\), we experimentally observed that \(C(d_1^i, d_2^i)\) has a zero-mean Normal distribution, i.e. \(E[C(d_1^i, d_2^i)]=0\). For all the experiments we run, we verified that the descriptors generated by our network are identically distributed and have zero-mean, therefore we impose \(l^i=0\) for non-matching samples.

With the training labels defined as indicated, the network can be trained via error gradient backpropagation, finding the parameters minimizing a loss function that, for the \textit{i}-th training sample, is defined as

\begin{equation}
L(l_i,y_i) = (l_i-C(d_1^i,d_2^i))^2.
\label{eqn:loss}
\end{equation}
\\*
That is, the network is trained to minimize a sample classification error function defined as the square error between the desired and the actual network outcome (i.e., the sample label). Regarding the practical aspects of the training procedure, we verified that normalizing the input patches over their own \(l_2\) norm increases the robustness and generalization capacity of the network with respect to illumination variations. Also, we follow the common practice of normalizing the input patches with respect to mean pixel intensity and standard deviation as calculated over the entire training set. We also observed that Spatial Batch Normalization as defined in \cite{ioffe2015batch} speeds up the training and boosts performance. Similarly, we normalize the DCT features \(F_D\) with respect to mean and standard deviation values over the entire coefficients set so that their value share the same distribution of the convolutional features \(F_C\).

\subsection{Descriptor Quantization}
\label{sec:quantization}

Binary descriptors enable lower bitrates and simple Hamming distance computation. Having verified that the trained network generates zero-mean descriptors, we quantize the B-elements descriptors \((d_1,d_2)\) over 1 bit via sign quantization, obtaining single bit, B-long descriptors \((b_1,b_2)\). Our choice of the cosine distance as similarity function is instrumental to comparing the binary descriptors \((b_1,b_2)\) via a simple Hamming distance. In fact, the cosine distance definition between \(d_1\), and \(d_2\) in Eq.~\ref{eqn:cosine} can be recasted as the L2-normalized inner product between \(d_1\) and \(d_2\). Given the definition of Hamming distance

\begin{equation}
H(b_1,b_2) \delequal \|b_1\| + \|b_2\| -2\langle b_1,b_2\rangle,
\label{eqn:hd}
\end{equation}
\noindent
where \(\langle \cdot,\cdot \rangle\) indicates the inner product, and \(\|\cdot\|\) refers to the bit count of the sequence, it follows that the cosine distance is equivalent to the Hamming distance, up to a scaling factor and an additive term. Therefore, pairs of binary descriptors are simply compared via the normalized Hamming distance

\begin{equation}
H_N(b_1,b_2) = \frac{1}{B} H(b_1,b_2),
\label{eqn:nhd}
\end{equation}
\noindent
which is close to 0 for pairs of matching patches, and close to 1 otherwise, and performance is computed as explained in the following section.
\section{Experimental Evaluation}
\label{sec:experiments}
\begin{figure}[t]
\centering
\includegraphics[width=4.5cm]{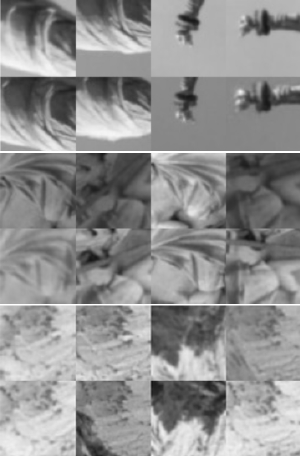}
\caption{Examples of matching pairs of patches (vertical) extracted from the Liberty (L), Notredame (N) and Yosemite (Y) datasets.}
\label{fig:lny}
\end{figure}

We evaluate our proposed architecture for patch matching over three datasets \cite{brown2011discriminative} of 64$\times$64 patches extracted from 3D reconstructions of the \textit{Liberty} statue (\(L\)), the \textit{Notredame} cathedral (\(N\)) and the \textit{Yosemite} mountains (\(Y\)), centered upon Difference of Gaussians (DoG) key points with canonical scale and orientation, as shown in Fig.~\ref{fig:lny}. Following the approach of \cite{brown2011discriminative}, we train the network three times, one for each dataset, and we evaluate its performance on the two other datasets for a total of 6 different training/testing \textit{setups} per experiment (e.g., the \(L/N\) setup indicates training on Liberty and testing on Notredame). For each setup, the network is trained with \(250000\) pairs of matching pairs and \(250000\) pairs of non-matching patches, whereas testing is performed over \(25000\) pairs of matching and \(25000\) pairs of non-matching patches. We implemented the proposed architecture using the \textit{Torch7} framework and trained our network following the procedure described in the Sec.~\ref{sec:training} over an NVIDIA K80 GPU. We rely on gradient descent with adaptive gradient optimization as defined in \cite{duchi2011adaptive}, with an initial learning rate set to \(10^{-4}\) and over batches of \(100\) pairs of samples (\(100\) matching samples and \(100\) non-matching samples per batch). The training ends when the error on the testing set has not decreased over the past \(10\) epochs or after \(400\) epochs. Coherently with \cite{brown2011discriminative}, first we compute ROC curves by thresholding the normalized Hamming distance between pairs of binary B-long descriptors, then, for each setup, we measure the False Positive Rate (FPR) for a True Positive Rate (TPR) set to \(95\)\%. In the following, we refer to the FPR computed for a TPR of \(95\)\% simply as \textit{patch classification error} for the sake of brevity.

Preliminarily, we experiment to find the convolutional configuration of the network architecture in Fig.~\ref{fig:struttura} that yields the best baseline performance when \(B=128\) bits. Namely, we want to find the best tradeoff between the number and resolution of convolutional \textit{featuremaps}. To this end, we first vary the number of convolutional modules \(M \in \{2,3,4\}\) (as M increases, the number of \textit{featuremaps} doubles while their resolution is halved horizontally and vertically). Furthermore, we experiment dropping the \textit{maxpooling} layer in \textit{M}-th convolutional module which additionally allows us to double the \textit{featuremaps} resolution without affecting their count. As this experiment focuses on convolutional features, we deactivate the DCT branch of each subnetwork (i.e., \(F_D=0\)). Table~\ref{tab:1} shows the performance of six different network configurations (\textit{-mp} rows account for the cases where the \textit{M}-th module \textit{maxpooling} layer is disabled). \(FMN\) and \(FMR\) columns indicate number and resolution of the \textit{featuremaps} respectively, where \(FMN \times FMR = F_C\). We observe that  \(8\times8\) \textit{featuremaps} consistently yield the best results: our hypothesis is that larger \textit{featuremaps} do not convey enough semantic information, whereas smaller \textit{featuremaps} lack the spatial resolution to preserve texture details.

\begin{table}[htpb]
\centering
 \centerline{
 \fontsize{6}{10}\selectfont{
\begin{tabular}{|c|c|c|c|c|c|c|c|c|c|}
    \hline
    \textbf{M} & \textbf{FMN} & \textbf{FMR} & \textbf{N/L} & \textbf{Y/L} & \textbf{L/N} & \textbf{Y/N} & \textbf{L/Y} & \textbf{N/Y} & \textbf{AVG} \\
    \hline
    \textbf{2} & \multirow{2}{*}{128} & 16x16 & 9.77 & 13.75 & 6.34 & 7.41 & 12.03 & 10.66 & 9.99 \\ 
    \textbf{2\textit{-mp}} & & 32x32 & 11.41 & 14.72 & 8.13 & 8.11 & 13.01 & 11.81 & 11.20 \\ 
    \hline
    \textbf{3} & \multirow{2}{*}{256} & 8x8 & 8.75 & 13.11 & 5.55 & 6.68 & 11.53 & 9.83 & \textbf{9.24} \\ 
    \textbf{3\textit{-mp}} & & 16x16 & 10.16 & 13.42 & 6.43 & 7.15 & 11.9 & 10.25 & 9.77 \\ 
    \hline
    \textbf{4} & \multirow{2}{*}{512} & 4x4 & 9.14 & 13.44 & 6.63 & 6.68 & 12.08 & 10.42 & 9.68 \\ 
    \textbf{4\textit{-mp}} & & 8x8 & 8.58 & 13.17 & 5.41 & 6.92 & 11.28 & 9.74 & \textbf{9.18} \\
    \hline
\end{tabular}
}
}
\vspace{5pt}
\caption{Patch classification error [\%] as a function of the number and resolution of convolutional featuremaps \((B=128)\)}
\label{tab:1}
\end{table}

\FloatBarrier

Next, we evaluate the entire network performance when DCT features are fused with convolutional features while accounting for network complexity. Input patches resolution is \(64\times64\) pixels, so each input patch DCT can be represented as a vector of \(4096\) real-valued coefficients. To keep the complexity low, we consider only the first \(F_D=561\) lower frequency coefficients selected as described in Sec.~\ref{sec:structure}. The additional network complexity due to the introduction of the DCT features, defined as \(F_D/F_C\), is negligible as it takes the values of 3.42\% and 1.71\% for the \(M=3\) and \(M=4\) cases, respectively. Table~\ref{tab:2} shows the results of the experiment: the second column reports the actual number of convolutional features \(F_C\) compared to the number of DCT features \(F_D=561\). As a general trend, we see that fusing DCT and convolutional features always improve the performance. Most importantly, the configuration with \(M=3\) convolutional modules (\(16945\) fused features) now yields better performance than the \(M=4-\textit{mp}\) configuration (\(33329\) fused features), despite the overall network complexity is reduced nearly by a twofold factor (about 9M parameters versus 17M parameters). The experiment shows that fusing convolutional features with transformed domain features further improves performance while keeping the network complexity under control.

\begin{table}[htbp]
\centering
 \centerline{\fontsize{6.7}{10}\selectfont{
\begin{tabular}{|c|c|c|c|c|c|c|c|c|}
    \hline
    \multirow{2}{*}{\textbf{M}}  & \boldmath $F_C$ \unboldmath & \multirow{2}{*}{\textbf{N/L}} & \multirow{2}{*}{\textbf{Y/L}} & \multirow{2}{*}{\textbf{L/N}} & \multirow{2}{*}{\textbf{Y/N}} & \multirow{2}{*}{\textbf{L/Y}} & \multirow{2}{*}{\textbf{N/Y}} & \multirow{2}{*}{\textbf{AVG}}\\
    & \boldmath $+ F_D$ \unboldmath & & & & & & & \\
    \hline
    \textbf{3} & 16348 & 8.75 & 13.11 & 5.55 & 6.68 & 11.53 & 9.83 & 9.24 \\ 
    \textbf{3,DCT} & +561 & 8.83 & 12.32 & 5.54 & 6.31 & 11.16 & 9.84 & \textbf{9.00} \\ 
    \hline
    \textbf{4\textit{-mp}} & 32768 & 8.58 & 13.17 & 5.41 & 6.92 & 11.28 & 9.74 & 9.18 \\ 
    \textbf{4\textit{-mp},DCT} & +561 & 9.11 & 12.22 & 5.85 & 6.30 & 11.33 & 9.98 & 9.13 \\ 
    \hline
    \textbf{4} & 8192 & 9.14 & 13.44 & 6.63 & 6.68 & 12.08 & 10.42 & 9.68 \\ 
    \textbf{4,DCT} & +561 & 9.03 & 12.99 & 6.00 & 6.55 & 12.10 & 10.35 & 9.50 \\
    \hline
\end{tabular}
}}
\vspace{5pt}
\caption{Patch classification error [\%] as a function of the type and number of features: feature fusion enables the best performance with controlled complexity increase \((B=128)\).}
\label{tab:2}
\end{table}

\begin{table}[b]
\centering
\begin{tabular}{|c|c|c|}
    \hline
        \ & \textbf{N/L} & \textbf{N/Y} \\ \hline
        \textbf{Common False Negatives [\%]} & 29.0 & 28.5 \\ 
        \textbf{Common False Positives [\%]} & 16.5 & 14.7 \\ 
    \hline
\end{tabular}
\vspace{5pt}
\caption{Fraction of patch pairs misclassified both by convolutional and DCT features: different types of features lead to different errors, thus are complementary \((B=128)\).}
\label{tab:3}
\end{table}

\begin{figure}[t]
\centering
\includegraphics[width=0.5\columnwidth]{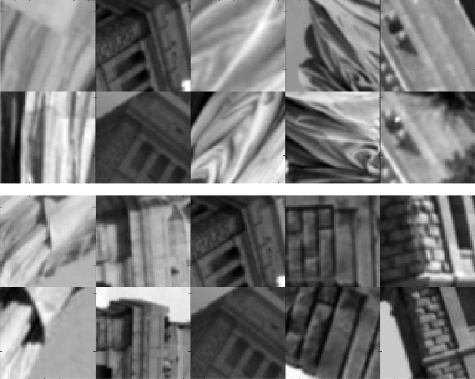}
\caption{Top-5 misclassified non-matching samples for the Notredame/Liberty (N/L) setup with convolutional features only (top row) and DCT features only (bottom row).}
\label{fig:top5}
\end{figure}

We attempt now to understand why feature fusion boosts performance. Without loss of generality, we repeat the previous experiment over a smaller training set of \(100000\) training pairs extracted from the Notredame dataset (\(50\)\% matching, \(50\)\% non matching), testing over \(10000\) pairs of patches from the Liberty (i.e. N/L setup) and \(10000\) pairs from the Yosemite (i.e. N/Y setup) datasets. We aim at assessing the effect of convolutional and DCT features separately, on an equal number of features basis (i.e., \(F_C=F_D\)). Therefore, we first experiment with a configuration of our architecture in Fig.~\ref{fig:struttura} where all the features generated by the DCT are provided as input to the fully connected layers without zig-zag selection, whereas the convolutional features are dropped altogether (i.e., \(F_D=4096, F_C=0\)). Next, we experiment dropping all the DCT features and considering instead a convolutional architecture with \(M=5\) modules (i.e., \(F_D=0, F_C=4096\)). Table~\ref{tab:3} reports the percentage of misclassified pairs of patches in both scenarios, quantified as the intersection of the sets over their union. We break down such erroneously classified pairs of patches in terms of \textit{common false negatives} and \textit{common false positives} for a threshold on the normalized Hamming Distance computed over the quantized descriptors that are equal to \(0.325\). The ratio of pairs of patches that are misclassified in both scenarios is significantly lower than \(30\)\%. This experiment suggests that convolutional and DCT features convey complementary information, thus explaining the increased ability to discern between patches of matching and non-matching pairs we observed when fusing convolutional and DCT features. Fig.~\ref{fig:top5} shows the Top-5 misclassified nonmatching samples (corresponding to a \(29\)\% of \textit{common false positives} in Table~\ref{tab:3}). Only the second column in the left image (convolutional features only) is also found as third column in the right image (DCT features only), confirming that different types of features yield to complementary classification errors.

\begin{figure}[t]
\centering
  \includegraphics[width=0.5\columnwidth]{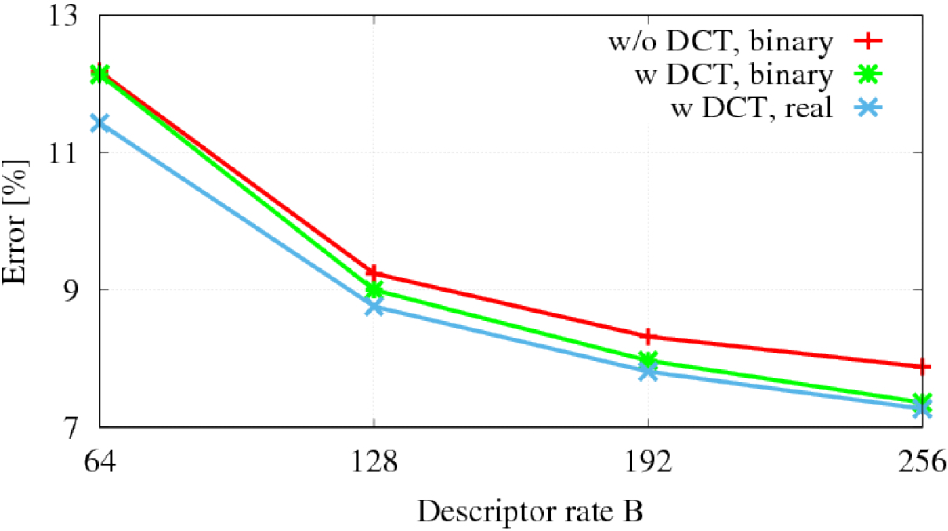}
\caption{The patch classification error decreases as B increases; this happens faster when applying feature fusion.}
\label{fig:patch_matching}
\end{figure}
\begin{table*}[t]
\centering
 \centerline{\fontsize{9}{12}\selectfont{
\begin{tabular}{|c|c|c|c|c|c|c|c|c|}
    \hline
        \textbf{~~~REF~~~} & \textbf{~~~B [bit]~~~} & \textbf{~~~N/L~~~} & \textbf{~~~Y/L~~~} & \textbf{~~~L/N~~~} & \textbf{~~~Y/N~~~} & \textbf{~~~L/Y~~~} & \textbf{~~~N/Y~~~} & \textbf{~~~AVG~~~} \\
        \hline
        \textbf{BinBoost} & \multirow{2}{*}{64} & \multirow{2}{*}{16.90} & \multirow{2}{*}{22.88} & \multirow{2}{*}{20.49} & \multirow{2}{*}{18.97} & \multirow{2}{*}{21.67} & \multirow{2}{*}{14.54} & \multirow{2}{*}{19.24}\\
        \textbf{\cite{trzcinski2013boosting}} & & & & & & & & \\
        \hline
        \textbf{DELFT} & 64 & 19.11 & 20.43 & 13.59 & 12.34 & 19.25 & 17.36 & 17.01\\
        \textbf{\cite{liu2016euclidean}} & 128 & 16.27 & 18.32 & 9.31 & 10.56 & 15.77 & 13.94 & 14.03\\
        \hline
        \textbf{Cov.Opt} & \multirow{2}{*}{1024} & \multirow{2}{*}{8.25} & \multirow{2}{*}{14.84} & \multirow{2}{*}{12.16} & \multirow{2}{*}{8.50} & \multirow{2}{*}{14.84} & \multirow{2}{*}{7.09} & \multirow{2}{*}{11.00}\\
        \textbf{\cite{simonyan2014learning}} & & & & & & & & \\
        \hline
        \textbf{DeepCD} & 192 & 8.31 & 16.29 & 14.45 & 13.62 & 18.38 & 8.97 & 13.34\\
        \textbf{\cite{yang2017deepcd}} & 768 & \textbf{3.73} & \textbf{9.97} & 7.82 & 7.67 & 11.75 & \textbf{4.35} & 7.55\\
        \hline
        \multirow{4}{*}{\textbf{Proposed}} & 64 & 11.68 & 16.48 & 8.52 & 9.13 & 14.16 & 12.89 & 12.14 \\
        & 128 & 8.83 & 12.32 & 5.54 & 6.31 & 11.16 & 9.84 & 9.00 \\
        & 192 & 7.56 & 11.74 & 4.84 & 5.40 & 9.71 & 8.58 & 7.97 \\
        & 256 & 7.30 & 10.87 & \textbf{4.29} & \textbf{4.83} & \textbf{8.83} & 8.05 & \textbf{7.36} \\
    \hline
\end{tabular}
}}
\vspace{5pt}
\caption{Patch classification error [\% FPR, TPR=95\%] of our framework compared with different  competitors: our approach consistently yields best results on an equal bit rate basis.}
\label{tab:4}
\end{table*}

Next, we experiment varying the descriptor bitrate beyond the \(B=128\) bits considered in our experiments so far. We experiment with the configuration \(M=3\) with feature fusion (\(F_C=16384,F_D=561\)), which provided the best performance-complexity trade-off  in Table~\ref{tab:2}, and its counterpart with convolutional features only (\(F_D=0\)). As expected, Fig.~\ref{fig:patch_matching} shows that the patch classification error decreases when B increases. Most importantly, Fig.~\ref{fig:patch_matching} shows that the error decreases faster when fusing features, i.e. feature fusion improves the performance-bitrate trade-off of our framework. Fig.~\ref{fig:patch_matching} includes a third line that corresponds to the configuration where non-quantized descriptors are considered in place of binary descriptors. For \(B=64\), the performance loss due to descriptor quantization is lower than 1\% and further decreases rapidly as B increases. Hence, sign quantization precisely matches the descriptors space into the Hamming space conserving the relative distance between vectors and allowing vector similarity preservation.

Finally, Table 4 compares our framework with several state-of-the-art approaches in binary patch matching. Our approach outperforms all the competitors on an equal bitrate basis. Our approach is outperformed on the N/L, Y/L, and N/Y setups only by \textit{DeepCD} \cite{yang2017deepcd}, which however relies on an aggregated descriptor rate of \(768\) bits that largely exceeds our maximum considered bitrate of \(B=256\) bits.
\section{Conclusions and future works}
\label{sec:conclusion}
In this work, we proposed to fuse convolutional features from the convolutional layers with features from the discrete cosine transform in a deep neural network for binary patch matching. Qualitative experiments suggest that different types of features are complementary in discriminating patches. Quantitative experiments over three challenging datasets confirm that our feature fusion approach outperforms several existing state of the art techniques based on convolutional features only. A careful design of the network topology and training procedures allowed us to capture distinctive features within the patches, enabling a very competitive architecture that also deploys feature fusion, allowing for performance improvement. We leave for our future investigations experimenting with different transform functions and evaluating our framework within a complete image matching pipeline.
\bibliographystyle{IEEEtran}
\bibliography{main}

\begin{thebibliography}{10}
\providecommand{\url}[1]{#1}
\csname url@samestyle\endcsname
\providecommand{\newblock}{\relax}
\providecommand{\bibinfo}[2]{#2}
\providecommand{\BIBentrySTDinterwordspacing}{\spaceskip=0pt\relax}
\providecommand{\BIBentryALTinterwordstretchfactor}{4}
\providecommand{\BIBentryALTinterwordspacing}{\spaceskip=\fontdimen2\font plus
\BIBentryALTinterwordstretchfactor\fontdimen3\font minus
  \fontdimen4\font\relax}
\providecommand{\BIBforeignlanguage}[2]{{%
\expandafter\ifx\csname l@#1\endcsname\relax
\typeout{** WARNING: IEEEtran.bst: No hyphenation pattern has been}%
\typeout{** loaded for the language `#1'. Using the pattern for}%
\typeout{** the default language instead.}%
\else
\language=\csname l@#1\endcsname
\fi
#2}}
\providecommand{\BIBdecl}{\relax}
\BIBdecl

\bibitem{zheng2017sift}
L.~Zheng, Y.~Yang, and Q.~Tian, ``Sift meets cnn: A decade survey of instance
  retrieval,'' \emph{IEEE Transactions on Pattern Analysis and Machine
  Intelligence}, 2017.

\bibitem{zagoruyko2015learning}
S.~Zagoruyko and N.~Komodakis, ``Learning to compare image patches via
  convolutional neural networks,'' in \emph{Proceedings of the IEEE Conference
  on Computer Vision and Pattern Recognition}, 2015, pp. 4353--4361.

\bibitem{simo2015discriminative}
E.~Simo-Serra, E.~Trulls, L.~Ferraz, I.~Kokkinos, P.~Fua, and F.~Moreno-Noguer,
  ``Discriminative learning of deep convolutional feature point descriptors,''
  in \emph{Proceedings of the IEEE International Conference on Computer
  Vision}, 2015, pp. 118--126.

\bibitem{zbontar2016stereo}
J.~Zbontar and Y.~LeCun, ``Stereo matching by training a convolutional neural
  network to compare image patches,'' \emph{Journal of Machine Learning
  Research}, vol.~17, no. 1-32, p.~2, 2016.

\bibitem{yang2017deepcd}
T.-Y. Yang, J.-H. Hsu, Y.-Y. Lin, and Y.-Y. Chuang, ``Deepcd: Learning deep
  complementary descriptors for patch representations,'' in \emph{Proceedings
  of the IEEE Conference on Computer Vision and Pattern Recognition}, 2017, pp.
  3314--3322.

\bibitem{zhu2016deep}
H.~Zhu, M.~Long, J.~Wang, and Y.~Cao, ``Deep hashing network for efficient
  similarity retrieval.'' in \emph{AAAI}, 2016, pp. 2415--2421.

\bibitem{yi2016lift}
K.~M. Yi, E.~Trulls, V.~Lepetit, and P.~Fua, ``Lift: Learned invariant feature
  transform,'' in \emph{European Conference on Computer Vision}.\hskip 1em plus
  0.5em minus 0.4em\relax Springer, 2016, pp. 467--483.

\bibitem{sorwar2001texture}
G.~Sorwar, A.~Abraham, and L.~S. Dooley, ``Texture classification based on dct
  and soft computing,'' in \emph{Fuzzy Systems, 2001. The 10th IEEE
  International Conference on}, vol.~2.\hskip 1em plus 0.5em minus 0.4em\relax
  IEEE, 2001, pp. 545--548.

\bibitem{pan2000image}
Z.~Pan, A.~G. Rust, and H.~Bolouri, ``Image redundancy reduction for neural
  network classification using discrete cosine transforms,'' in \emph{Neural
  Networks, 2000. IJCNN 2000, Proceedings of the IEEE-INNS-ENNS International
  Joint Conference on}, vol.~3.\hskip 1em plus 0.5em minus 0.4em\relax IEEE,
  2000, pp. 149--154.

\bibitem{chopra2005learning}
S.~Chopra, R.~Hadsell, and Y.~LeCun, ``Learning a similarity metric
  discriminatively, with application to face verification,'' in \emph{Computer
  Vision and Pattern Recognition, 2005. CVPR 2005. IEEE Computer Society
  Conference on}, vol.~1.\hskip 1em plus 0.5em minus 0.4em\relax IEEE, 2005,
  pp. 539--546.

\bibitem{bertinetto2016fully}
L.~Bertinetto, J.~Valmadre, J.~F. Henriques, A.~Vedaldi, and P.~H. Torr,
  ``Fully-convolutional siamese networks for object tracking,'' in
  \emph{European Conference on Computer Vision}.\hskip 1em plus 0.5em minus
  0.4em\relax Springer, 2016, pp. 850--865.

\bibitem{ioffe2015batch}
S.~Ioffe and C.~Szegedy, ``Batch normalization: Accelerating deep network
  training by reducing internal covariate shift,'' in \emph{International
  Conference on Machine Learning}, 2015, pp. 448--456.

\bibitem{brown2011discriminative}
M.~Brown, G.~Hua, and S.~Winder, ``Discriminative learning of local image
  descriptors,'' \emph{IEEE transactions on pattern analysis and machine
  intelligence}, vol.~33, no.~1, pp. 43--57, 2011.

\bibitem{duchi2011adaptive}
J.~Duchi, E.~Hazan, and Y.~Singer, ``Adaptive subgradient methods for online
  learning and stochastic optimization,'' \emph{Journal of Machine Learning
  Research}, vol.~12, no. Jul, pp. 2121--2159, 2011.

\bibitem{trzcinski2013boosting}
T.~Trzcinski, M.~Christoudias, P.~Fua, and V.~Lepetit, ``Boosting binary
  keypoint descriptors,'' in \emph{Proceedings of the IEEE Conference on
  Computer Vision and Pattern Recognition}, 2013, pp. 2874--2881.

\bibitem{liu2016euclidean}
Z.~Liu, Z.~Li, J.~Zhang, and L.~Liu, ``Euclidean and hamming embedding for
  image patch description with convolutional networks,'' in \emph{Proceedings
  of the IEEE Conference on Computer Vision and Pattern Recognition Workshops},
  2016, pp. 72--78.

\bibitem{simonyan2014learning}
K.~Simonyan, A.~Vedaldi, and A.~Zisserman, ``Learning local feature descriptors
  using convex optimisation,'' \emph{IEEE Transactions on Pattern Analysis and
  Machine Intelligence}, vol.~36, no.~8, pp. 1573--1585, 2014.

\end{thebibliography}
\end{document}